\documentclass[12pt,journal,compsoc,onecolumn]{IEEEtran}
\usepackage{amsmath,amssymb,amsfonts,graphicx}
\usepackage{graphicx}
\usepackage{multirow}
\usepackage{url}

\usepackage{dsfont}
\usepackage{pifont}
\usepackage[noadjust]{cite}
\usepackage{nicematrix}
\let\OLDthebibliography\thebibliography
\renewcommand\thebibliography[1]{
  \OLDthebibliography{#1}
  \setlength{\parskip}{0pt}
  \setlength{\itemsep}{1.47pt plus 0.3ex}
}

\def\SS{{\cal S}}
\def\G{{\cal G}}
\def\V{{\cal V}}
\def\E{{\cal E}}
\def\A{{\bf A}}
\def\F{{\cal F}}
\def\W{{\bf W}}

\def\UU{{\bf U}} 
\def\N{{\cal N}}
\def\WW{{\bf W}}
\def\M{{\bf M}}

\title{Training Lightweight Graph Convolutional Networks with Phase-field Models}
\author{\IEEEauthorblockN{Hichem Sahbi \\ $ $ \\}
\IEEEauthorblockA{Sorbonne University, UPMC, CNRS, LIP6, F-75005 Paris, France}}

\begin{document}

\maketitle

\begin{abstract}
In this paper, we design lightweight graph convolutional networks (GCNs) using a particular class of regularizers,  dubbed as phase-field models (PFMs).  PFMs exhibit a bi-phase behavior using a particular ultra-local term that allows training both the topology and the weight parameters of GCNs as a part of a single ``end-to-end'' optimization problem.  Our proposed solution  also relies on a reparametrization that pushes the mask of the topology towards binary values leading to effective topology selection and high generalization while implementing any targeted pruning rate.  Both masks and weights share the same set of latent variables and this further enhances the generalization power of the resulting lightweight GCNs.   Extensive experiments conducted on the challenging task of skeleton-based recognition show the outperformance of PFMs against other staple regularizers as well as related lightweight design methods.
\end{abstract}
  
\section{Introduction}

Deep convolutional networks are nowadays becoming mainstream in solving many pattern classification tasks  including image and action recognition \cite{Krizhevsky2012}.   Their principle consists in training convolutional filters together with pooling and attention  mechanisms that maximize classification performances. Many existing  convolutional networks were initially dedicated to grid-like data, including images \cite{He2016,Girshick15}.  However,  data sitting on top of irregular domains (such as skeleton graphs  in action recognition) require extending    convolutional networks  to  general graph structures, and these extensions are known as graph convolutional networks (GCNs) \cite{Bruna2013,Henaff2015}.  Two families of GCNs exist in the literature: spectral and spatial. The former achieve convolutions using Fourier ~\cite{kipf17,Levie2018,Li2018,Bresson16} whilst the latter are based on message passing, via attention matrices, prior to convolution \cite{Gori2005,Micheli2009,Wu2019,Hamilton2017}. Whereas spatial GCNs have been relatively more effective compared to spectral ones, their precision is highly reliant on the accuracy of the attention matrices that capture context and node-to-node relationships \cite{Bengio17}. With multi-head attention, GCNs are more accurate but overparametrized and computationally overwhelming.\\
\indent Many solutions are proposed in the literature in order to reduce  time and  memory  footprint of  convolutional networks including GCNs.   Some of them pretrain oversized networks prior to reduce their computational complexity  (using distillation \cite{DBLP:journals/corr/HintonVD15,DBLP:conf/iclr/ZagoruykoK17,DBLP:journals/corr/RomeroBKCGB14,DBLP:conf/aaai/MirzadehFLLMG20,DBLP:conf/cvpr/ZhangXHL18,DBLP:conf/cvpr/AhnHDLD19}, tensor decomposition~\cite{Denton2014,Wang2018},  quantization~\cite{DBLP:journals/corr/HanMD15,Courbariaux16,Gupta15,Chen15,Rastegari16,Park17} and pruning \cite{Tung18,DBLP:conf/iclr/0022KDSG17,DBLP:conf/nips/CunDS89,DBLP:conf/nips/HassibiS92,DBLP:conf/nips/HanPTD15,DBLP:conf/iccv/LiuLSHYZ17}), while others build efficient networks from scratch using neural architecture search  \cite{Zoph17}.  In particular, pruning methods, either unstructured or structured are  currently  becoming  mainstream. Their principle consists in removing connections whose impact on the classification performance is the least noticeable. Structured pruning \cite{DBLP:conf/iclr/0022KDSG17,DBLP:conf/iccv/LiuLSHYZ17,Zhao19} consists in removing groups of connections,  entire filters, etc., and this makes the class of learnable subnetworks highly rigid. In contrast, unstructured pruning \cite{DBLP:conf/nips/HanPTD15,DBLP:journals/corr/HanMD15} is more flexible and proceeds by dropping-out connections individually using different proxy criteria,  such as weight magnitude~\cite{DBLP:conf/nips/HanPTD15} or using more sophisticated variational methods~\cite{Durk15,David18,Mol17}.\\ 
\noindent The general recipe of variational pruning consists in learning both the weights and the binary masks that capture the topology of the pruned subnetworks. This is achieved by minimizing an objective function that combines (via a mixing hyperparameter) a classification loss and a regularizer which controls the sparsity of the resulting masks \cite{DBLP:conf/iccv/LiuLSHYZ17,Wen16,Louizos18}. However, these methods are powerless to implement any given targeted pruning rate (cost) without overtrying multiple settings of the mixing hyperparameters. Alternative variational methods model explicitly the cost, using $\ell_0$-based criteria \cite{Louizos18,Pan16}, in order to minimize the discrepancy between the observed cost and the targeted one. Nonetheless, the underlying optimization problems are highly combinatorial and existing solutions usually rely on sampling heuristics. Existing more tractable relaxation (such as $\ell_1$/$\ell_2$-based, etc.~\cite{Gordon18,Miguel18,Lemaire2019}) promote sparsity, but are powerless to implement any given target cost {\it exactly}, and also result into overpruning effects leading to disconnected subnetworks, with weak generalization, especially at very high pruning regimes. Besides, most of the existing pruning solutions decouple the training of network topology (masks) from weights, and this doubles the number of training parameters and increases the risk of overfitting. Finally, the mainstream magnitude pruning \cite{DBLP:conf/nips/HanPTD15} allows reaching any targeted cost, but relies on a tedious fine-tuning step and also decouples the training of topology from weights and this weakens generalization.\\

\indent Considering all these issues,  we introduce in this paper a new lightweight network design based on the phase-field model (PFM). The latter gathers the upsides of the aforementioned regularization methods while discarding their downsides at some extent. PFM is based on an ultra-local term with two local minima around $0$ and $1$; when composed with a particular mask reparametrization,  PFM promotes sparsity by pushing the values of this reparametrization towards crisp (binary) values without any {\it ill-posed} annealing mechanism. In other words, the proposed method allows generating only feasible solutions (i.e., binary masks) while implementing any targeted pruning rate without overtrying multiple mixing hyperparameters. The proposed solution also avoids the decoupling of weights and masks, and this reduces the number of training parameters. Experiments conducted on the challenging task of action and hand-gesture recognition show a consistent gain of the proposed PFM-based approach against staple regularizers and cost-sensitive variational methods as well as the related work including magnitude pruning.

\section{A Glimpse on GCNs}
Let $\SS=\{\G_i=(\V_i, \E_i)\}_i$ denote a collection of graphs with $\V_i$, $\E_i$ being respectively the nodes and the edges of $\G_i$. Each graph $\G_i$ (denoted for short as $\G=(\V, \E)$) is endowed with a signal $\{\phi(u) \in \mathbb{R}^s: \ u \in \V\}$ and associated with an adjacency matrix $\A$. GCNs aim at learning a set of $C$ filters $\F$ that define convolution on $n$ nodes of $\G$ (with $n=|\V|$) as
 \begin{equation}\label{matrixform}
(\G \star \F)_\V = f\big(\A \  \UU^\top  \   \W\big),
 \end{equation}
 \noindent  here $^\top$ stands for transpose,  $\UU \in \mathbb{R}^{s\times n}$  is the  graph signal, $\W \in \mathbb{R}^{s \times C}$  is the matrix of convolutional parameters corresponding to the $C$ filters and  $f(.)$ is a nonlinear activation applied entry-wise. In Eq.~(\ref{matrixform}), the input signal $\UU$ is projected using $\A$ and this provides for each node $u$, the  aggregate set of its neighbors. Entries of $\A$ could be handcrafted or learned so Eq.~(\ref{matrixform}) implements a convolutional block with two layers; the first one aggregates signals in $\N(\V)$ (sets of node neighbors) by multiplying $\UU$ with $\A$ while the second layer achieves convolution by multiplying the resulting aggregates with the $C$ filters in $\W$. Learning  multiple adjacency (also referred to as attention) matrices (denoted as $\{\A^k\}_{k=1}^K$) allows us to capture different contexts and graph topologies when achieving aggregation and convolution.  With multiple matrices $\{\A^k\}_k$ (and associated convolutional filter parameters $\{\W^k\}_k$),  Eq.~(\ref{matrixform}) is updated as  $(\G \star \F)_\V = f\big(\sum_{k=1}^K \A^k   \UU^\top     \W^k\big)$. Stacking aggregation and convolutional layers, with multiple  matrices $\{\A^k\}_k$, makes GCNs accurate but heavy. We propose, in what follows, a method that makes our networks lightweight and still effective.
 \section{Lightweight Design}

In the rest of this paper,   a given GCN is  subsumed as a multi-layered neural network $g_\theta$  whose weights defined as $\theta =  \left\{\WW^1,\dots, \WW^L \right\}$, with $L$ being its depth,  $\WW^\ell \in \mathbb{R}^{d_{\ell-1} \times d_{\ell}}$ its $\ell^\textrm{th}$  layer weight tensor, and $d_\ell$ the dimension of $\ell$. The output of a given layer  $\ell$ is defined as
$ \mathbf{\phi}^{\ell} = f_\ell({\WW^\ell}^\top \  \mathbf{\phi}^{\ell-1})$, $\ell \in \{2,\dots,L\}$,  being $f_\ell$ an activation function; without a loss of generality, we omit the bias in the definition of  $\mathbf{\phi}^{\ell}$. Pruning consists in zeroing-out a subset of weights in $\theta$ by multiplying $\WW^\ell$ with a binary mask $\M^\ell \in \{ 0,1 \}^{d_{\ell-1} \times d_{\ell}}$. The binary entries of  $\M^\ell$ are set depending on whether the underlying layer connections are kept or removed, so $\mathbf{\phi}^{\ell} = f_\ell( (\M^\ell \odot \WW^\ell )^\top \ \mathbf{\phi}^{\ell-1} )$, here $\odot$ stands for the element-wise matrix product. In this definition, entries of the tensor $\{\M^\ell\}_\ell$ are set depending on the prominence of the underlying connections in $g_\theta$. However, such pruning suffers from several drawbacks. On the one hand, optimizing the discrete set of variable $\{\M^\ell\}_\ell$ is known to be highly combinatorial and intractable especially on large networks. On the other hand,  the total number of parameters $\{\M^\ell\}_\ell$, $\{\WW^\ell\}_\ell$ is twice the number of connections in $g_\theta$ and this increases training complexity  and may also lead to overfitting. In order to circumvent these issues, we consider an alternative {\it reparametrization} that allows finding both the topology of the pruned networks together with their weights, without doubling the size of the training parameters, while making learning still effective.
\subsection{ Weight Reparametrization} We consider an alternative parametrization of the network related to magnitude pruning. This reparametrization corresponds to the Hadamard product involving a weight tensor and a function applied entry-wise to the same tensor as
\begin{eqnarray}\label{eq2}
\WW^\ell = \hat{\WW}^\ell \odot \psi(\hat{\WW}^\ell).
\end{eqnarray}

\noindent In the above equation, $\hat{\WW}^\ell$ is a latent tensor and $\psi(\hat{\WW}^\ell)$ is a continuous relaxation of $\M^\ell$ which enforces the prior that smallest weights should be removed from the network. In order to achieve this goal,  $\psi$ must be (i) bounded in $[0,1]$, (ii) differentiable, (iii) symmetric, and (iv) $\psi(\omega) \leadsto 1$ when $|\omega|$ is sufficiently large and $\psi(\omega) \leadsto 0$ otherwise. The first and the fourth properties ensure that the reparametrization is neither acting as a scaling factor greater than one nor changing the sign of the latent weight, and also acts as the identity for sufficiently large weights, and as a contraction factor for small ones. The second property is necessary to ensure that $\psi$ has computable gradient  while the third condition guarantees that only the magnitudes of the latent weights matter\footnote{A possible choice, used in practice, that satisfies these four conditions (when combined with PFM) is $\psi(\omega)=2\sigma(\omega^2)-1$ with $\sigma$ being the sigmoid function.}. Note that the fourth property is implemented without any ill-posed temperature annealing, but instead using a  phase-field model  --  presented subsequently  -- which  controls the smoothness of $\psi$ around the support of the latent weights. Put differently,  the asymptotic behavior of $\psi$   -- that allows selecting the topology of the pruned subnetworks  -- is obtained using the phase-field energy as described below.

\subsection{Phase-field Model}
\begin{figure}[h]
  \resizebox{1.0\columnwidth}{!}{\centering
   \includegraphics[width=0.3\linewidth]{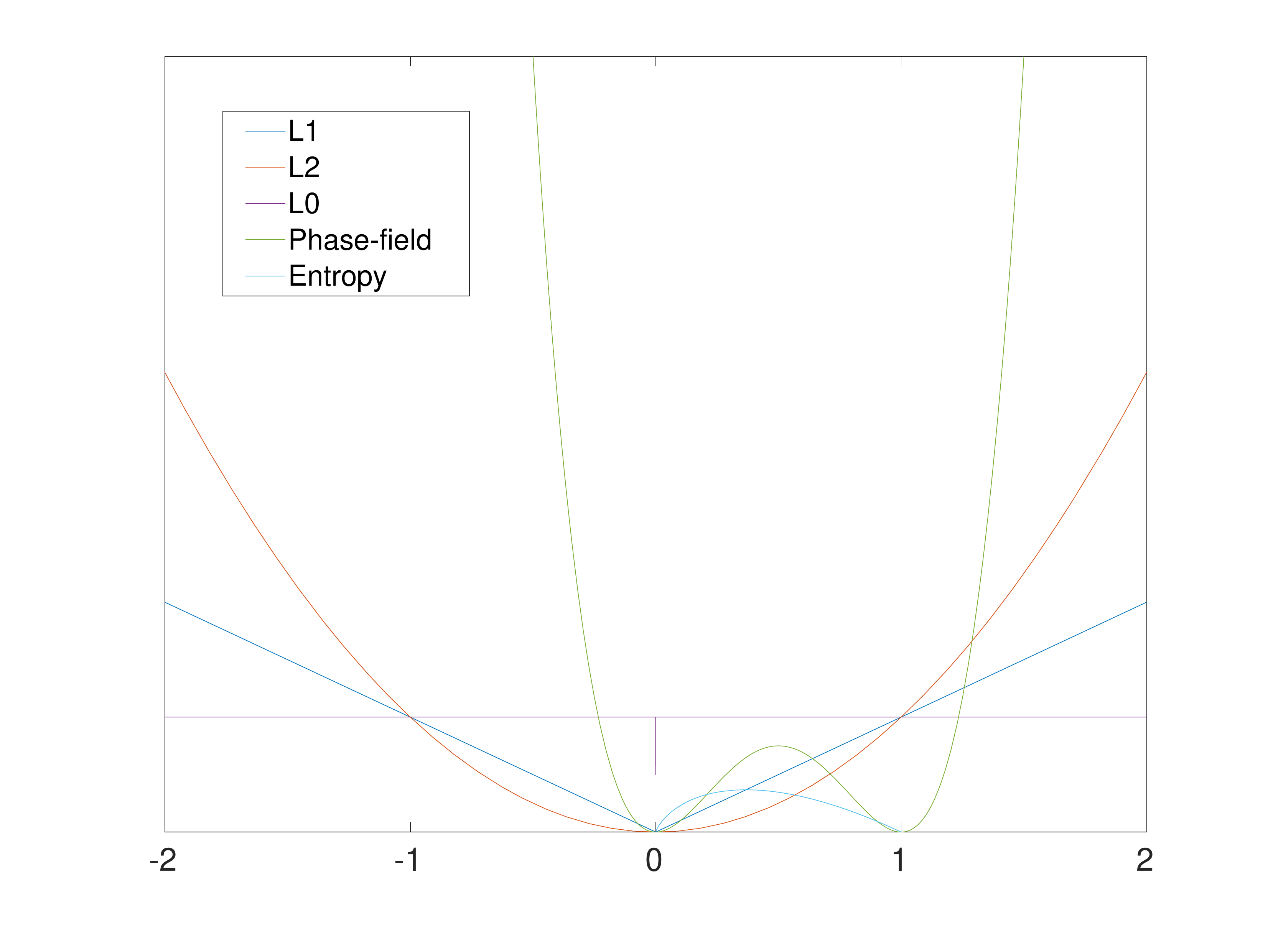}\hspace{-0.25cm}\includegraphics[width=0.3\linewidth]{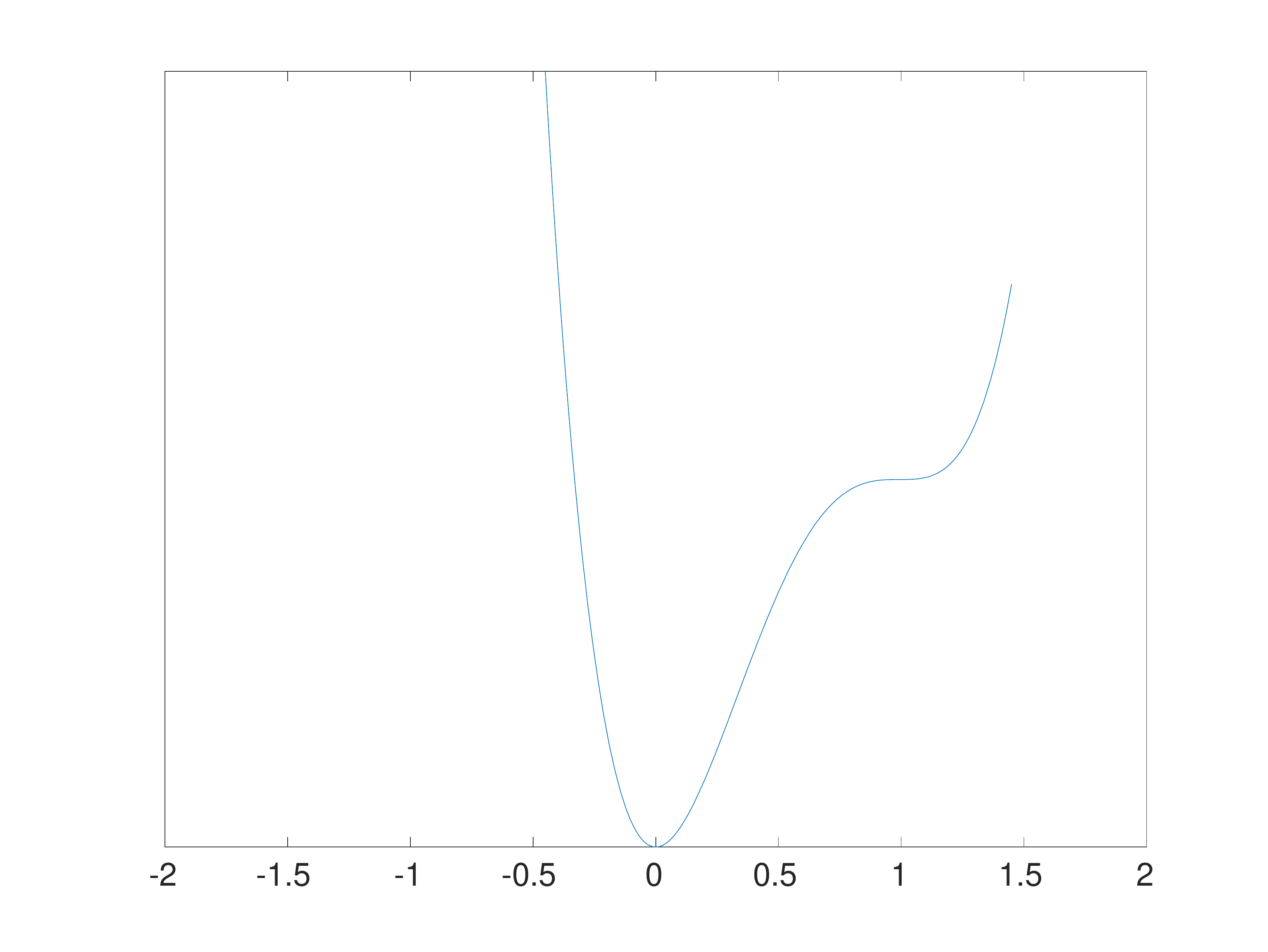}\hspace{-0.25cm}\includegraphics[width=0.3\linewidth]{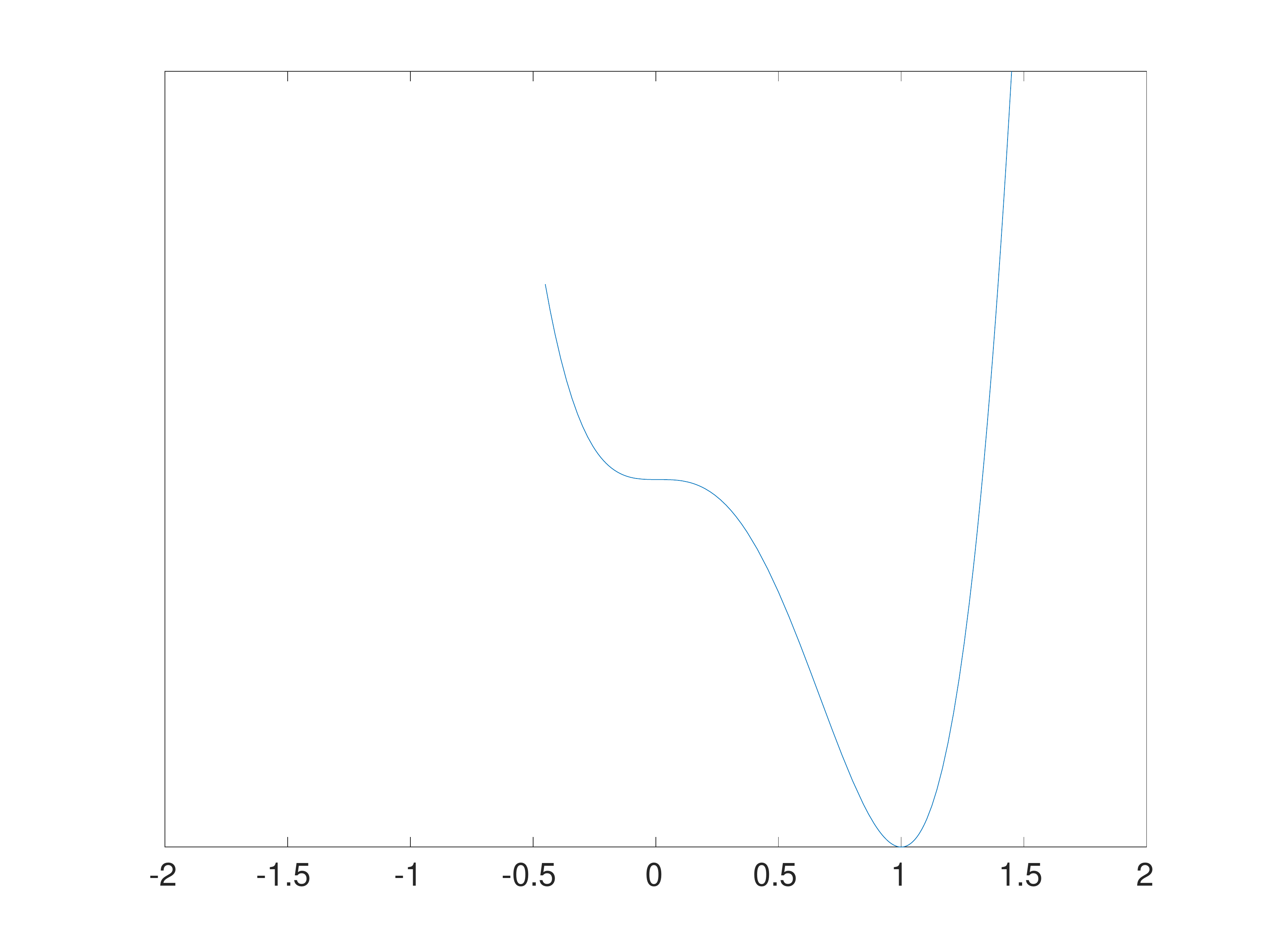}}
\caption{(Left) This figure shows a comparison of different regularizers, namely $\ell_0$, $\ell_1$, $\ell_2$ and entropy as well as the balanced  PFM (i.e., $\alpha=0$). From these curves, it is clear that our PFM gathers the advantages of all these regularizers: (i) it strongly penalizes large mask values (as $\ell_2$), (ii) it (not only) pushes mask values near 0 (as all the regularizers), but also near 1 (as $\ell_0$ and entropy) which allows counting non-zeros while (iii) behaving as a relaxed differentiable function (as $\ell_1$), (iv) finally, similarly to entropy, PFM has also two local minima between $0$ and $1$, however entropy does not penalize large mask values and does not allow reaching any (a priori) fixed pruning rate. (Mid and right) These figures show the two imbalanced versions of PFM (corresponding to $\alpha>0$ and $\alpha<0$) that allow implementing over and under-pruning respectively. In all experiments, $\beta$ is arbitrarily fixed to $3$ while $\alpha$ is accordingly chosen depending on the targeted pruning rate $\textrm{tpr}=\frac{\alpha+\beta}{2\beta}$ (which corresponds to the local maximum of ultra-local term).}\label{tab20}
\end{figure}

A phase-field is a real-valued function defined on an input domain $\Omega \subset \mathbb{R}$ \cite{sahbi2014}. A phase-field determines a region by the map $\xi_z(\psi)=\{\omega \in \Omega: \psi(\omega) > z \}$ (where $z$ is a given threshold) and a  phase-field energy as
\begin{eqnarray}\label{eq00}
   E_P(\psi) = \displaystyle  \int_{\Omega} \ \displaystyle V(\psi(\omega))  \ d\omega,
\end{eqnarray}
\noindent here $V(.)$  - referred to as the ultra-local term  - is given by
\begin{eqnarray}
\beta \bigg( \frac{(2t-1)^4}{4}- \frac{(2t-1)^2}{2} \bigg)  + \alpha \bigg(2t-1-\frac{(2t-1)^3}{3}\bigg).
\end{eqnarray}
If one minimizes (\ref{eq00}) subject to $\xi_z (\psi) = {\cal R}$ for a fixed region ${\cal R}$, then away from the boundary $\partial {\cal R}$, the minimizing function (denoted as $\psi_{{\cal R}}$) assumes approximately the value +1 inside, and 0 outside ${\cal R}$ thanks to the ultra-local term, and it varies smoothly (depending on $\beta$) across the interface near $\partial {\cal R}$. Considering a discrete approximation of the integral on a finite set of parameters $\{\omega_i\}_i$, one may rewrite $E_P(\psi)$ as
$E_P(\psi(\{\omega_i\}_i)) = \sum_i  V(\psi(\omega_i))$.
In order to guarantee two energy minima at 0 and +1 associated to the two classes (pruned/unpruned), the inequality $\beta > |\alpha|$ must be satisfied, so $V'(1) = V'(0)= 0$ and $V''(1)=V''(0) > 0$ where $'$ and $''$ denote the first and second derivatives respectively. Notice that the formulation of our PFM yields to choose the threshold $z$ to be at the maximum $(\beta+\alpha)/2\beta$ of $V$ which also corresponds to the fixed pruning rate (see Fig.~\ref{tab20}). When setting $\alpha = 0$ and $\beta >0$, we get the particular case of PFM; this leads to $V (1) = V (0)$ corresponding to equiprobable phases $\{0,1\}$, and hence  $E_P$ is suitable for balanced pruning. In contrast, for significantly imbalanced pruning (which is the main scope of this paper), one should select $\alpha \neq 0$  so that the two phases $\{0,1\}$ will have different energies: $V(1)-V(0) = 4 \alpha/3 \neq 0$. It is clear that the sign of $\alpha$ affects the behavior of $E_P$: if $\alpha > 0$ then the model prefers to reduce the volume of $\cal R$ and vice-versa (see again Fig.~\ref{tab20}). This behavior corresponds to the core contribution of our work that allows achieving significantly imbalanced pruning.

\subsection{Variational Pruning}
Pruning is achieved using a global loss as a combination of a cross-entropy term ${\cal L}_e$, and the phase-field energy  $E_P$ (which controls the cost and aims at zeroing as much mask entries as possible depending on the setting of $\alpha$) resulting into
\begin{equation}\label{eq3}
  \begin{array}{ll}
  \displaystyle  \min_{\{\hat{\WW}^\ell\}_\ell}  \displaystyle {\cal L}_e\big(\{\hat{\WW}^\ell \odot \psi(\hat{\WW}^\ell)\}_\ell\big)  \ + \ \lambda \ E_P(\{\psi(\hat{\WW}^\ell)\}_\ell), 
\end{array}
    \end{equation}
    \noindent    here $\lambda$ is sufficiently large (overestimated in practice), so Eq.~(\ref{eq3}) focuses  on binarizing  $\{\psi(\hat{\WW}^\ell)\}_\ell$ using the phase-field energy, and also constraining the pruning rate to reach $\frac{\alpha+\beta}{2\beta}$. As training evolves, $E_P$  reaches its minimum and stabilizes while the gradient of the global loss becomes dominated by the gradient of ${\cal L}_e$, and this maximizes further the classification performances. 
\section{Experiments}
In this section, we evaluate the performances of our pruned GCNs on skeleton-based recognition using two challenging datasets, namely SBU~\cite{Kiwon12} and FPHA~\cite{Garcia2018}. SBU is an interaction dataset acquired using the Microsoft Kinect sensor; it includes in total 282 moving skeleton sequences (performed by two interacting individuals) belonging to 8 categories. Each pair of interacting individuals corresponds to two 15 joint skeletons and each joint is encoded with a sequence of its 3D coordinates across video frames. In this dataset, we consider the same evaluation protocol as the one suggested in the original dataset release~\cite{Kiwon12} (i.e., train-test split).  The FPHA dataset includes 1175 skeletons  belonging to 45 action categories with high inter and intra subject variability.  Each skeleton includes 21 hand joints and each joint is again encoded with a sequence of its 3D coordinates across video frames. We evaluate the performance of our method following the protocol in~\cite{Garcia2018}. In all these experiments, we report the average accuracy over all the classes of actions.\\

\noindent {\bf Implementation details.} We trained the GCNs end-to-end using the Adam optimizer \cite{Adam2014} for 2,700 epochs  with a batch size equal to $200$ for SBU and $600$ for FPHA, a momentum of $0.9$ and a global learning rate (denoted as $\nu(t)$)  inversely proportional to the speed of change of the loss used to train our networks. When this speed increases (resp. decreases),   $\nu(t)$  decreases as $\nu(t) \leftarrow \nu(t-1) \times 0.99$ (resp. increases as $\nu(t) \leftarrow \nu(t-1) \slash 0.99$). In all these experiments, we use a GeForce GTX 1070 GPU (with 8 GB memory). The architecture of our baseline GCN (similar to \cite{Sahbi2022}) includes an attention layer of 1 head on SBU (resp. 16 heads on FPHA) applied to skeleton graphs whose nodes are encoded with 8-channels (resp. 32 for FPHA), followed by a convolutional layer of 32 filters for SBU (resp. 128 filters for FPHA),  and a dense fully connected layer and a softmax layer.
The initial network for SBU is not heavy, its number of parameters does not exceed 15,320, and this makes its pruning challenging as many connections will be isolated (not contributing in the evaluation of the network output). In contrast, the initial network for FPHA is relatively heavy (for a GCN) and its number of parameters reaches 2 millions. As shown subsequently, both GCNs are accurate compared to the related work on the SBU/FPHA benchmarks~\cite{Sahbi2020,Sahbi2022}. Considering these GCN baselines, our goal is to make them highly lightweight while making their accuracy as high as possible. \\

\begin{figure}[h]
\centering \includegraphics[width=0.5\linewidth]{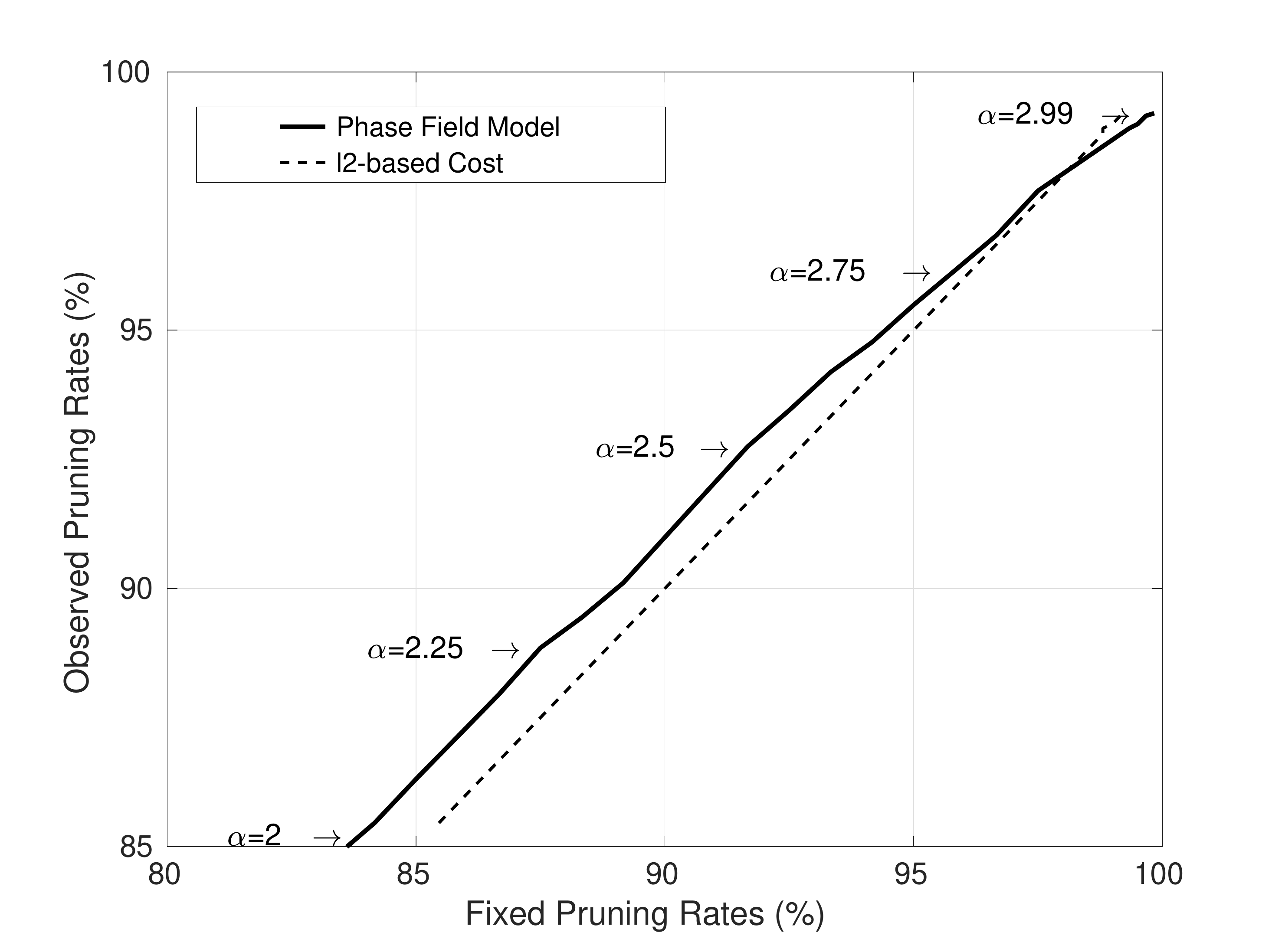}\includegraphics[width=0.5\linewidth]{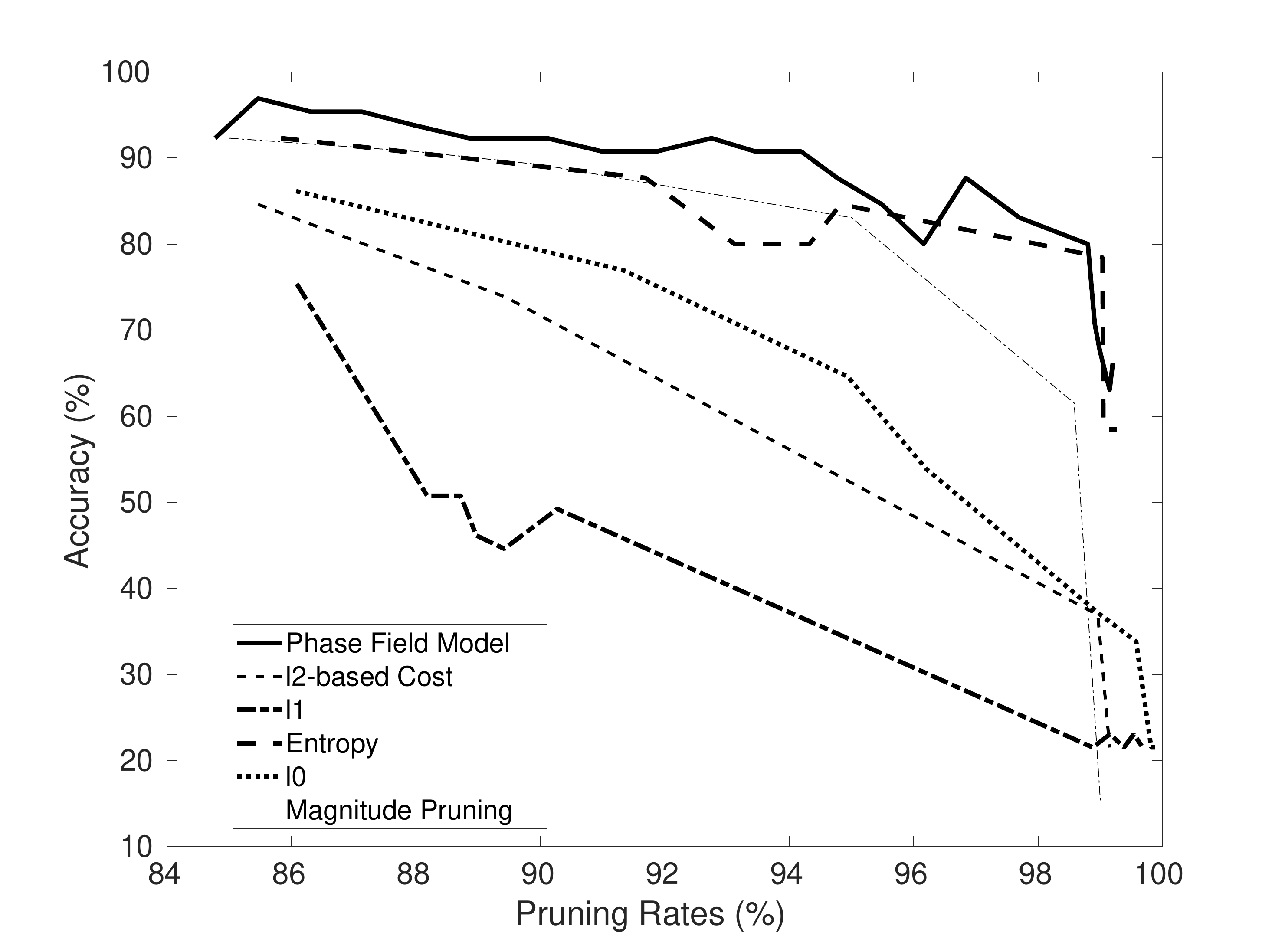}
 \caption{(Left) This figure shows the alignment between the fixed and the observed pruning rates when using cost-sensitive pruning and PFMs. (Right) The underlying accuracy w.r.t. different pruning rates. Note that $\beta=3$ in all the PFMs.}\label{tab2}\end{figure}

\noindent {\bf Model analysis and comparison.} Fig~\ref{tab2}-left shows the alignment between the fixed/targeted pruning rates (tpr) and the observed rates when using PFM and its comparison against cost-sensitive pruning. In these experiments, PFM acts not only as a regularizer (and binarizer) but also as a rebalancing function which allows implementing any tpr by choosing $\alpha$ that satisfies $\frac{\alpha+\beta}{2\beta}=\textrm{tpr}$ or equivalently $\alpha=2\beta\times \textrm{tpr}-\beta$. Fig.~\ref{tab2}-right shows the accuracy of our lightweight GCNs w.r.t. the underlying pruning rates. In these results, PFM is compared against different {\it alternative} regularizers (plugged in Eq.~\ref{eq3} instead of PFM), namely $\ell_0$ \cite{Louizos18}, $\ell_1$ \cite{Koneru2019}, entropy \cite{Wiedemann2019} and $\ell_2$-based cost-sensitive pruning~\cite{Sahbi2022}. From these results, the impact of PFM is substantial on highly pruned GCNs while relatively smaller pruning regimes provide equivalent performances. Note that when alternative regularizers are used, multiple settings (trials) of the underlying hyperparameter $\lambda$ (in Eq.~\ref{eq3}) are considered prior to reach any given targeted pruning rate, and this makes the whole training and pruning process overwhelming. While cost-sensitive pruning makes training more tractable, its downside resides in the collapse of trained masks, and this degrades performances significantly at high pruning rates; a similar behavior is observed with magnitude pruning (see again Fig.~\ref{tab2}-right). \\
\indent Table~\ref{tab6} shows an ablation study (and extra comparisons) of our PFM when used individually and jointly with the other regularizers as well as cost-sensitive pruning. From these results, we first observe that when training is achieved with weight reparametrization, performances are equivalent and sometimes overtake the initial heavy GCN, with less parameters (pruning rate does not exceed 70\% as no control on tpr is achieved) as this produces a regularization effect similar to ~\cite{dropconnect2013}. Second, we observe a positive impact of PFM when jointly combined with the aforementioned regularizers and cost-sensitive loss; note that when PFM is jointly used, $\alpha$ is set to $0$, so $\textrm{tpr}=0.5$ and this makes the rebalancing effect of PFM null, and only the other regularizers allow implementing the targeted pruning rates when $\lambda$ is appropriately tuned. Finally, extra comparison against magnitude pruning~\cite{DBLP:journals/corr/HanMD15} shows the substantial gain of our PFM at very high pruning regimes. 

\begin{table}[ht]
 \begin{center}
\resizebox{0.99\columnwidth}{!}{
  \begin{tabular}{c|cccc}
  \rotatebox{0}{Datasets}   &   \rotatebox{0}{Methods}  & \rotatebox{0}{Pruning rates (\%)} &  \rotatebox{0}{\# Parameters} &  \rotatebox{0}{Accuracy (\%)}  \\ 
    \hline
    \hline
\multirow{13}{*}{ \rotatebox{30}{SBU}}  &  Initial Model~\cite{Sahbi2022}                       & 0.0        &  15320     &  90.76  \\ 
  &  Weight Reparametrization (WR)                 & 70.66 & 4494 & \bf93.84 \\

                            &  Magnitude Pruning~\cite{DBLP:journals/corr/HanMD15}     &   98.58         & 216  & 61.53  \\ 
    \cline{2-5}
  &  WR+$\ell_0$                             &       99.00     &   152      & 36.92   \\ 
                            &  WR+$\ell_0$+PFM ($\alpha=0$)                      &  99.05        &   144      &  \bf55.38  \\ 
        \cline{2-5}

  &  WR+$\ell_1$                              &    98.87        &  171        &   21.53  \\ 
                            &  WR+$\ell_1$+PFM ($\alpha=0$)                     &   98.94         &   161      & \bf73.84   \\ 
        \cline{2-5}

  &  WR+Entropy                       &      98.97     &     157    &    60.00 \\
                            & WR+Entropy+PFM ($\alpha=0$)                 &  98.96        &    158     &  \bf61.53  \\ 
        \cline{2-5}

   & WR+$\ell_2$-based Cost                       &         98.96          &   158      & 36.92 \\ 

   & WR+PFM  ($\alpha=2\beta \textrm{tpr}-\beta$)                          &  98.98  &  154  & 67.69  \\ 
   & WR+$\ell_2$-based Cost+PFM ($\alpha=0$)                 &  98.96         &   158      &   \bf75.38  \\ 
    \hline
    \hline
 \multirow{13}{*}{ \rotatebox{30}{FPHA}}  & Initial Model~\cite{Sahbi2022}                       & 0.0        &  1967616     &  86.08  \\ 
   & Weight Reparametrization (WR)                  & 50.38 &976268 & 85.56 \\ 
   & Magnitude Pruning~\cite{DBLP:journals/corr/HanMD15}     &    98.83        &22892   & 52.69  \\ 
        \cline{2-5}

 & WR+$\ell_0$                           &     99.24     &   14858     &  8.34  \\ 

                            & WR+$\ell_0$+PFM ($\alpha=0$)                      &  99.43        &  11203      &  \bf64.69  \\ 
        \cline{2-5}

   & WR+$\ell_1$                          &     99.26         &     14460    &  2.78  \\ 
                            & WR+$\ell_1$+PFM ($\alpha=0$)                     &  99.26         &    14460    &  \bf70.78 \\ 
       \cline{2-5}

   & WR+Entropy                          &   99.09        &   17788      & 31.13  \\
                            & WR+Entropy+PFM ($\alpha=0$)                 & 99.25        &   14683    & \bf67.47  \\ 
        \cline{2-5}

   & WR+$\ell_2$-based Cost                         & 99.49          &    9945  &   5.56      \\ 
   & WR+PFM  ($\alpha=2\beta \textrm{tpr}-\beta$)                        &  99.68         & 6156   & 65.91 \\ 
   & WR+$\ell_2$-based Cost+PFM ($\alpha=0$)                 &  99.49         &  10034      &   \bf69.91  \\ 
    \hline
  \end{tabular}
}
\end{center}
\caption{Ablation study of our pruning method (with and without PFMs). When PFMs are combined with other regularizers, $\alpha$ is necessarily equal to 0, so only the regularization effect is considered (as the other regularizers  indirectly control the pruning rate). When PFMs are individually used, $\alpha=2\beta \textrm{tpr}-\beta$ where tpr corresponds to the targeted pruning rate. In these results, PFMs are also compared against weight reparametrization and magnitude pruning. It's worth noticing that low accuracies result from the disconnected pruned networks obtained at very high pruning regimes.}\label{tab6}
\end{table}

\section{Conclusion}  
In this paper, we introduce a novel pruning method based on phase-field models (PFMs) which allow training very lightweight GCNs at very high pruning regimes. The strength of PFMs resides in their ability to leverage the advantage of different regularizers used in variational pruning while discarding their inconveniences at some extent. Indeed, the proposed PFMs allow training highly overpruned networks, binarizing the underlying masks while implementing any targeted pruning rate and improving generalization. Extensive experiments conducted on the challenging task of skeleton-based recognition show the substantial gain of our pruned lightweight networks against different baselines as well as the related work. As a future work, we are currently investigating the extension of this method to other network architectures and datasets.

\end{document}